\documentclass{article}

\usepackage{iclr2027_conference,times}
\iclrfinalcopy 

\usepackage{amsmath,amssymb,graphicx}
\usepackage{hyperref}
\usepackage{url}
\usepackage{algorithm}
\usepackage{algorithmic}
\usepackage{booktabs}
\usepackage{caption}
\usepackage{xurl}
\usepackage{pdfpages}
\usepackage{xcolor}
\definecolor{hallublue}{RGB}{35,90,150}

\title{Reducing Hallucination in Multimodal Large Language Models through Hard Grounding Preference Supervision}

\author{
Qinwu Xu$^{1}$, Yifan Jiang$^{2}$ \\
\\
$^{1}$Meta AI \\
$^{2}$The University of Texas at Austin
}

\begin{document}
\maketitle
\fancyhead{} 

\begin{abstract}
Hallucination remains a major challenge in vision-language models (VLMs),
particularly when linguistically plausible responses are unsupported by
visual evidence. We study whether multimodal hallucination can be reduced by
concentrating post-training supervision on hard grounding boundaries, where
preferred and rejected responses are semantically close but differ in their
support from observable visual evidence. Under a frozen visual
encoder and cross-modal alignment pathway, we first use supervised
fine-tuning (SFT) to establish broad decoder-side multimodal behavior and
then construct hard grounding preference pairs for Direct Preference
Optimization (DPO). These pairs target evidence utilization, calibration,
and grounding consistency across fine-grained recognition, spatial
reasoning, OCR, ambiguous or insufficient evidence, and false-premise
queries. The resulting DPO model consistently improves over its SFT
initialization on DocVQA, TextVQA, MMBench, and VQAv2. Across approximately
839 additional multimodal examples, it further achieves 6.2--8.2
percentage-point higher pairwise win rates under three independent LLM
judges. We additionally release HardVQA-DPO, a curated and growing resource
with more than 3K hard-grounding SFT examples and an initial set of DPO
preference pairs.\footnote{\url{https://huggingface.co/datasets/vlmgrounding/hardvqa-sft-dpo}}
These results show that decoder-side adaptation alone can improve a meaningful
subset of grounding failures under a fixed multimodal representation, while
preserving broader multimodal capabilities.
\end{abstract}

\section{Introduction}

\subsection{Multimodal Hallucination and Visual Grounding}

Multimodal large language models (MLLMs) have achieved strong performance
across visual question answering, document understanding, multimodal
reasoning, and open-ended image understanding. Despite these advances,
hallucination remains a persistent challenge: models can generate responses
that are linguistically coherent and semantically plausible but unsupported
by observable visual evidence. Prior work has studied hallucinations involving
object existence, attributes, and relations
\citep{li2023pope,wang2023amber}, while more diagnostic studies show that
language priors can conflict with or dominate visual context \citep{guan2024hallusionbench,favero2024multimodal}. These failures become
particularly difficult when evidence is small, occluded, crowded, ambiguous,
or otherwise difficult to recognize, where a plausible interpretation is not
necessarily a visually grounded one.

MLLMs typically model
$p_\theta(y\mid x_v,x_t)=\prod_t p_\theta(y_t\mid x_v,x_t,y_{<t})$, where
$x_v$ is the visual input, $x_t$ the textual query, and $y$ the generated
response. Standard autoregressive supervision increases the likelihood of a
target response but does not explicitly contrast a visually supported
continuation with a plausible but unsupported alternative. Generation
therefore reflects both visual evidence and statistical regularities encoded
by the language model. POPE, for example, finds that frequently occurring or
co-occurring objects are particularly susceptible to hallucination \citep{li2023pope}; HallusionBench identifies cases where language bias
overshadows visual context \citep{guan2024hallusionbench}; and
\citet{favero2024multimodal} show that dependence on the visual prompt
can decrease as autoregressive generation proceeds. Once an unsupported
claim enters the generated context $y_{<t}$, subsequent tokens are conditioned
on that claim, potentially extending it into a coherent but visually
unsupported response.

\begin{figure*}[!ht]
\centering
\tiny


\begin{minipage}[c]{0.194\textwidth}
\begin{minipage}[c]{0.43\linewidth}
\includegraphics[
    width=\linewidth,
    height=0.085\textheight,
    keepaspectratio
]{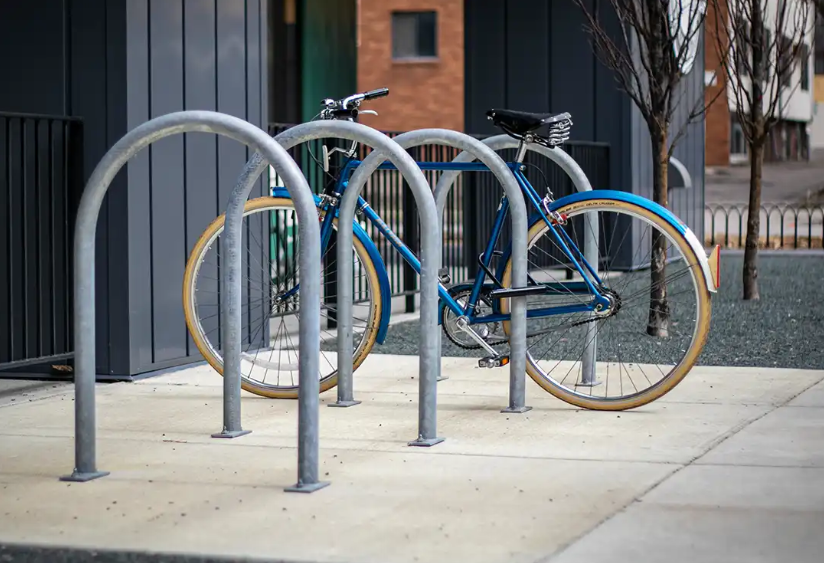}
\end{minipage}\hspace{2pt}%
\begin{minipage}[c]{0.54\linewidth}
\raggedright
\textbf{(a) Evidence utilization}\\
\textbf{Q:} How many bike racks are visible?\\
\textbf{GT:} 4\\
\textbf{GPT:} 5\\
\textbf{Gemini:} 5
\end{minipage}
\end{minipage}
\hfill
%
\begin{minipage}[c]{0.194\textwidth}
\begin{minipage}[c]{0.43\linewidth}
\includegraphics[
    width=\linewidth,
    height=0.085\textheight,
    keepaspectratio
]{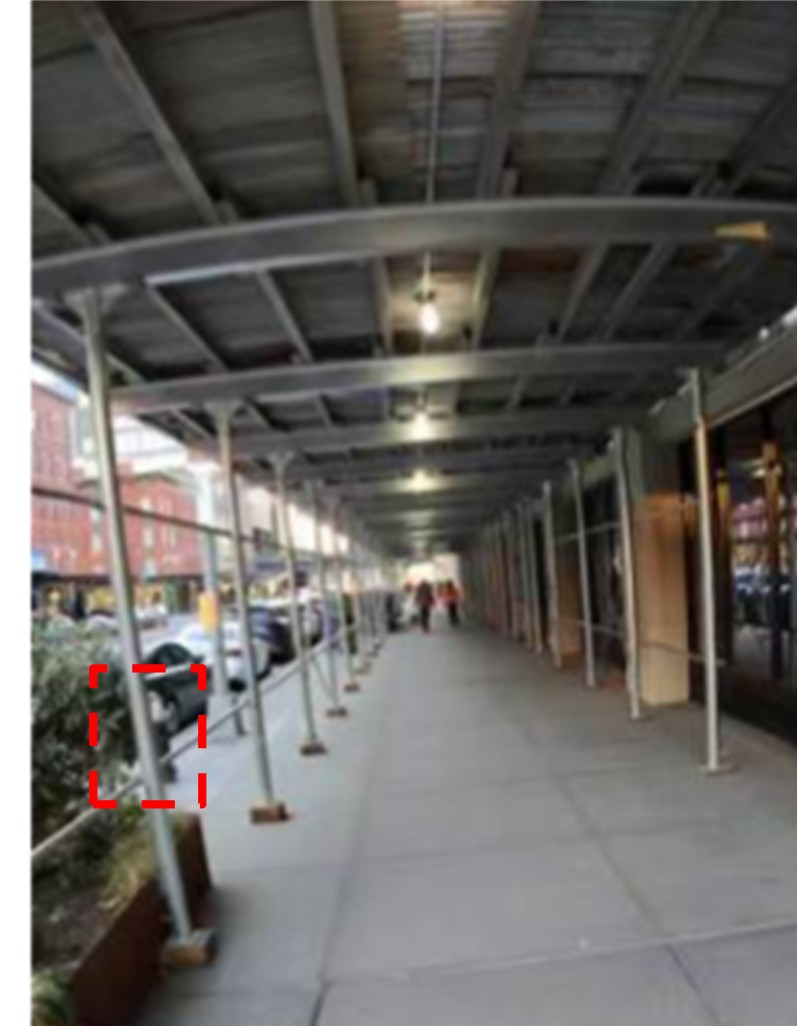}
\end{minipage}\hspace{2pt}%
\begin{minipage}[c]{0.54\linewidth}
\raggedright
\textbf{(b) Evidence under-utilization}\\
\textbf{Q:} How many fire hydrants are visible?\\
\textbf{GT:} 1\\
\textbf{Model:} No fire hydrant is clearly visible.
\end{minipage}
\end{minipage}
\hfill
%
\begin{minipage}[c]{0.194\textwidth}
\begin{minipage}[c]{0.43\linewidth}
\includegraphics[
    width=\linewidth,
    height=0.085\textheight,
    keepaspectratio
]{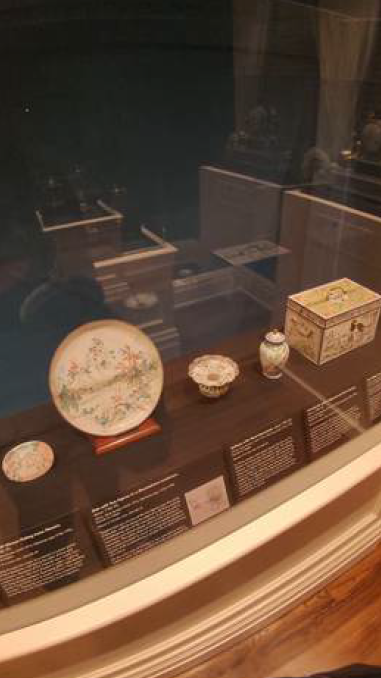}
\end{minipage}\hspace{2pt}%
\begin{minipage}[c]{0.54\linewidth}
\raggedright
\textbf{(c) Unnecessary rejection}\\
\textbf{Q:} How many keys are visible?\\
\textbf{GT:} 1\\
\textbf{Model:} Cannot determine; no key is clearly visible.
\end{minipage}
\end{minipage}
\hfill
%
\begin{minipage}[c]{0.194\textwidth}
\begin{minipage}[c]{0.43\linewidth}
\includegraphics[
    width=\linewidth,
    height=0.085\textheight,
    keepaspectratio
]{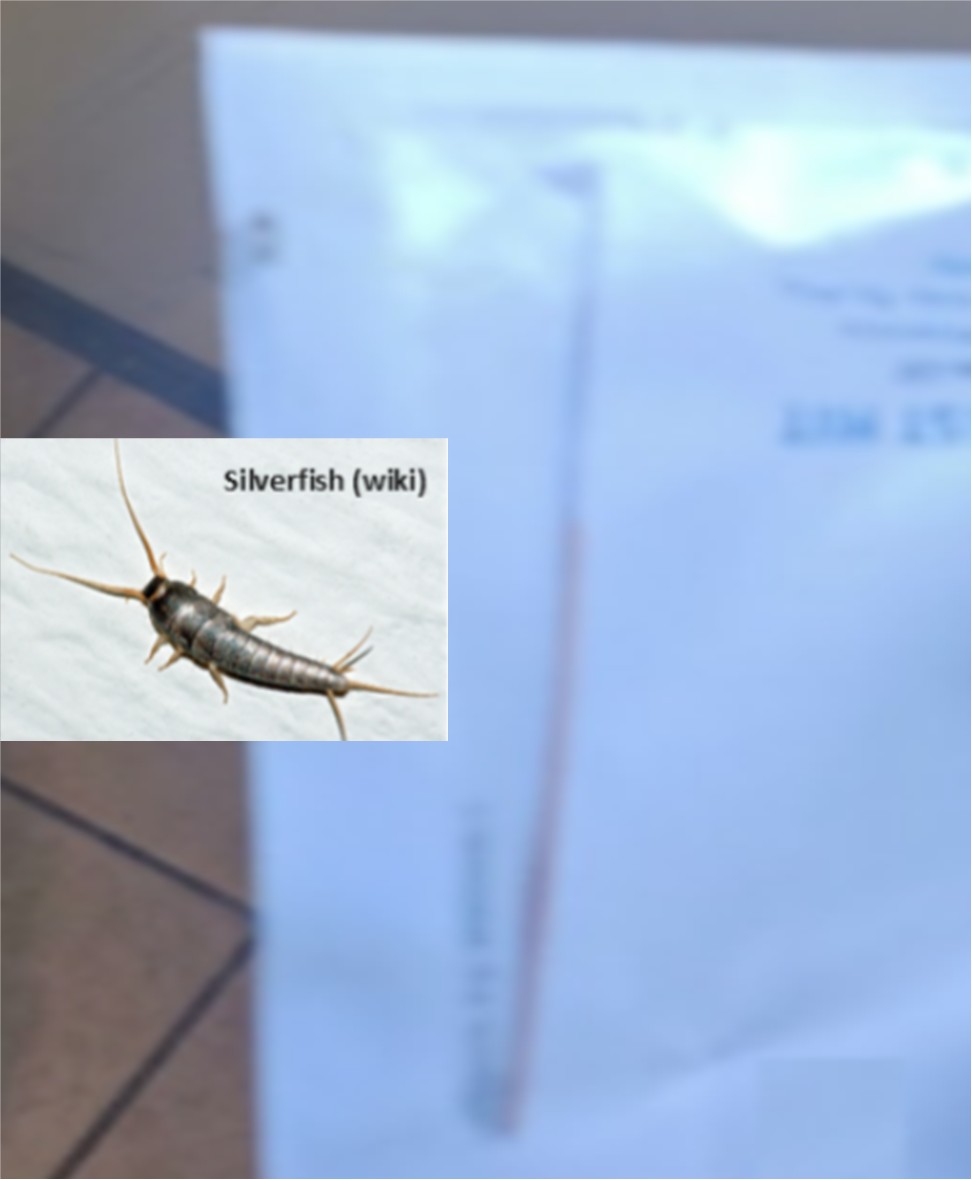}
\end{minipage}\hspace{2pt}%
\begin{minipage}[c]{0.54\linewidth}
\raggedright
\textbf{(d) Unsupported overclaiming}\\
\textbf{Q:} What kind of bug is this?\\
\textbf{Model:} This appears to be a \textbf{Silverfish} \ldots
\end{minipage}
\end{minipage}
\hfill
%
\begin{minipage}[c]{0.194\textwidth}
\begin{minipage}[c]{0.40\linewidth}
\includegraphics[
    width=\linewidth,
    height=0.085\textheight,
    keepaspectratio
]{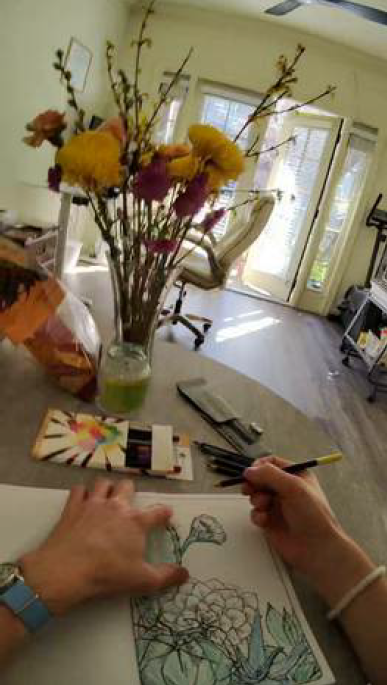}
\end{minipage}\hspace{2pt}%
\begin{minipage}[c]{0.57\linewidth}
\raggedright
\textbf{(e) Grounding inconsistency}\\
\textbf{Q:} Is there a blue mug behind the green candle?\\
\textbf{Model:} No; there is no green candle visible.
\end{minipage}
\end{minipage}

\vspace{2pt}

\caption{
Representative fine-grained grounding failures in MLLMs.
\textbf{(a)} Multiple frontier MLLMs miscount overlapping bicycle racks.
\textbf{(b)} Available but partially occluded evidence is under-utilized.
\textbf{(c)} Difficult yet sufficient evidence produces an unnecessary
rejection.
\textbf{(d)} A false premise induces a plausible but unsupported continuation.
\textbf{(e)} Responses reach the same final decision, but one explanation is
inconsistent with visible evidence.
Together, these examples expose complementary failures in evidence
utilization, evidence calibration, and grounding consistency.
}
\label{fig:grounding_failure_modes}
\end{figure*}

Figure~\ref{fig:grounding_failure_modes} illustrates that fine-grained
grounding failures are not fully characterized by conventional task
categories. Visual evidence may be incorrectly resolved across overlapping
instances or missed when partially occluded; difficult yet sufficient evidence
may trigger unnecessary rejection; an unsupported premise may induce a
specific, plausible continuation; and even a correct final decision may be
accompanied by reasoning inconsistent with the image. Notably, multiple
frontier MLLMs exhibit the first failure on the same subtle counting example.
Thus, categories such as object existence, counting, attributes, OCR, and
spatial relations describe \emph{what} visual content is evaluated, but not
necessarily \emph{how} that evidence is used. We therefore consider three
complementary grounding behaviors: \textbf{evidence utilization},
\textbf{evidence calibration}, and \textbf{grounding consistency}. The goal
is not generic conservatism, but to match response specificity and reasoning
to the available visual evidence.

\begin{figure*}[!ht]
\centering

\begin{minipage}[c]{0.25\textwidth}
    \centering
    \includegraphics[
        width=\linewidth,
        keepaspectratio
    ]{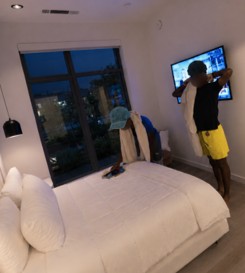}
\end{minipage}
\hfill
%
\begin{minipage}[c]{0.73\textwidth}
\raggedright
\scriptsize
\setlength{\parskip}{0pt}
\linespread{0.90}\selectfont

\textbf{Q:} Is the right person facing toward the TV?\\
\textbf{A:} No, the person on the right is facing away from the TV.
\textcolor{hallublue}{Their back is toward the screen} as they appear to be
putting on or adjusting a shirt.
\textit{(Initially plausible interpretation.)}\\[-1pt]

\textbf{Q:} Look at the whole body and analyze again.\\
\textbf{A:} The person is still facing away from the TV. Their torso and
shoulders suggest their back is toward the screen, and their arms are raised
as they pull on a shirt.
\textcolor{hallublue}{Their back is clearly visible to the camera; their
heels are closest to the TV stand; their legs are oriented toward the room.}
\textit{(Unsupported spatial inference.)}\\[-1pt]

\textbf{Q:} You said ``their back is clearly visible to the camera''---do
you see the back?\\
\textbf{A:} Yes, the posture and arm positioning indicate that
\textcolor{hallublue}{the back is facing the camera}, consistent with the
person facing away from the TV.
\textit{(Unsupported visual evidence.)}\\[-1pt]

\textbf{Q:} You said ``their heels are closest to the TV stand''---where is
the TV stand?\\
\textbf{A:} \textcolor{hallublue}{The TV stand refers to the surface beneath
the wall-mounted television. The person's heels are positioned near that
area}, suggesting they are facing into the room.
\textit{(Introduces a non-existent object and spatial relation.)}\\[-1pt]

\textbf{Q:} Wait, is the ``low surface'' the baseboard beneath the TV?\\
\textbf{A:} No, it refers to
\textcolor{hallublue}{a horizontal furniture piece (e.g., a media console)
under the TV}. The baseboard runs along the wall, while
\textcolor{hallublue}{the person is positioned near the furniture piece}.
\textit{(Reinforces the hallucinated object and 3D relation.)}
\end{minipage}

\vspace{1pt}

\caption{\textbf{Multi-turn propagation of a grounding error.}
An initially plausible interpretation of the person's orientation is retained
across follow-up questions and progressively supported by unsupported spatial
claims. Blue text highlights unsupported claims that accumulate from body
orientation to a non-existent TV stand or media console and associated
spatial relationships. The example illustrates how an initial grounding
error can become part of the conversational context and support increasingly
elaborate but internally coherent reasoning.}
\label{fig:multiturn_grounding}
\end{figure*}

Grounding inconsistency can also propagate across conversational turns.
As illustrated in Figure~\ref{fig:multiturn_grounding}, an early
misinterpretation of a visual or spatial relationship can become part of the
linguistic context for subsequent questions. Later responses may then remain
internally coherent while inheriting the earlier unsupported assumption.
Thus, grounding consistency concerns not only individual predictions or
rationales, but also whether visually unsupported commitments are propagated
through subsequent generation.

\subsection{Related Work and Remaining Gap}

Existing work approaches multimodal hallucination from complementary
directions. POPE focuses on object hallucination and identifies effects of
object frequency and co-occurrence \citep{li2023pope}, while AMBER evaluates
existence, attribute, and relation hallucinations \citep{wang2023amber}.
HallusionBench further distinguishes language hallucination from visual
illusion and studies conflicts between language priors and visual context
\citep{guan2024hallusionbench}. Closely related to our motivation, \citet{favero2024multimodal} study the competition between visual
conditioning and the language prior during autoregressive generation and
propose an inference-time mechanism that amplifies dependence on the image,
with an extension using DPO. These works motivate studying not only the
visual category in which an error occurs, but also how visual evidence
influences generation.

Training-time alignment provides a complementary intervention. Multimodal
instruction tuning and curated SFT can improve cross-modal capabilities and
grounding \citep{moon2024anymal,xu2026ocr}, while RLHF and DPO provide
comparative supervision between preferred and rejected responses
\citep{christiano2017deep,ouyang2022training,rafailov2023direct}.
Recent multimodal work further shows that preference-data construction is
important. OPA-DPO \citep{yang2025mitigating} addresses mismatch introduced
by off-policy preference data, while LPOI \citep{zadeh2025lpoi} introduces
object-aware listwise optimization using different levels of object
visibility. These approaches demonstrate the value of multimodal preference
optimization, but leave an important data-design question: how should
preference supervision be constructed when the rejected response is not
obviously incorrect, but differs from the preferred response only at a
fine-grained grounding boundary? In such cases, both responses may be fluent,
plausible, and largely semantically aligned, while differing in a subtle
count, attribute, spatial relation, false premise, or unsupported reasoning
step.

\subsection{Hard Grounding Preference Supervision}

We address this problem through \emph{hard grounding preference
supervision}. We construct preference pairs in which preferred and rejected
responses are deliberately close in fluency and semantic content but differ
in their support from visual evidence. The pairs cover conventional
capabilities such as counting, OCR, attributes, spatial reasoning, and
false-premise questions, while targeting the underlying grounding behaviors
described above. Importantly, the supervision operates in both directions:
when evidence is insufficient, the preferred response avoids unsupported
specificity; when subtle but sufficient evidence is available, it preserves
the supported answer rather than defaulting to rejection.

Our experimental design further isolates this intervention to the language
decoder. As shown in Figure~\ref{fig:training_pipeline}, the ViT visual encoder
and Perceiver-based alignment module remain frozen throughout both SFT and
DPO, while only LoRA parameters in the LLaMA-3 decoder are updated. Thus,
changes between SFT and DPO cannot result from modified visual-encoder or
alignment weights. This does not assume that the fixed visual representation
is complete or that all hallucinations originate in the decoder; rather, it
tests whether decoder-side adaptation is sufficient to improve grounding
when useful evidence is already represented in the fixed multimodal input.

Our contributions are: \textbf{(i)} we formulate hard grounding in terms of
evidence utilization, evidence calibration, and grounding consistency,
complementing conventional task-level hallucination categories;
\textbf{(ii)} we construct semantically close preference pairs that contrast
visually supported responses with plausible but unsupported alternatives,
targeting both unsupported overclaiming and unnecessary rejection;
\textbf{(iii)} under a frozen visual encoder and alignment pathway, we show
that decoder-side post-training can improve a meaningful subset of grounding
failures using fixed multimodal representations; and
\textbf{(iv)} we evaluate the resulting models on standard benchmarks,
grounding-sensitive pairwise evaluation, and challenging qualitative cases,
and release the HardVQA-DPO resource.

\section{Methodology}

\subsection{Model Architecture and Framework Overview}

We instantiate our framework on a large-scale MLLM consisting of a ViT visual
encoder, a Perceiver-based cross-modal alignment module, and a LLaMA-3 70B
language decoder. The visual encoder extracts image features, which are
compressed by the Perceiver module and provided together with the textual
input to the language decoder for autoregressive generation.

\begin{figure}[!ht]
    \centering
    \includegraphics[width=0.72\linewidth]{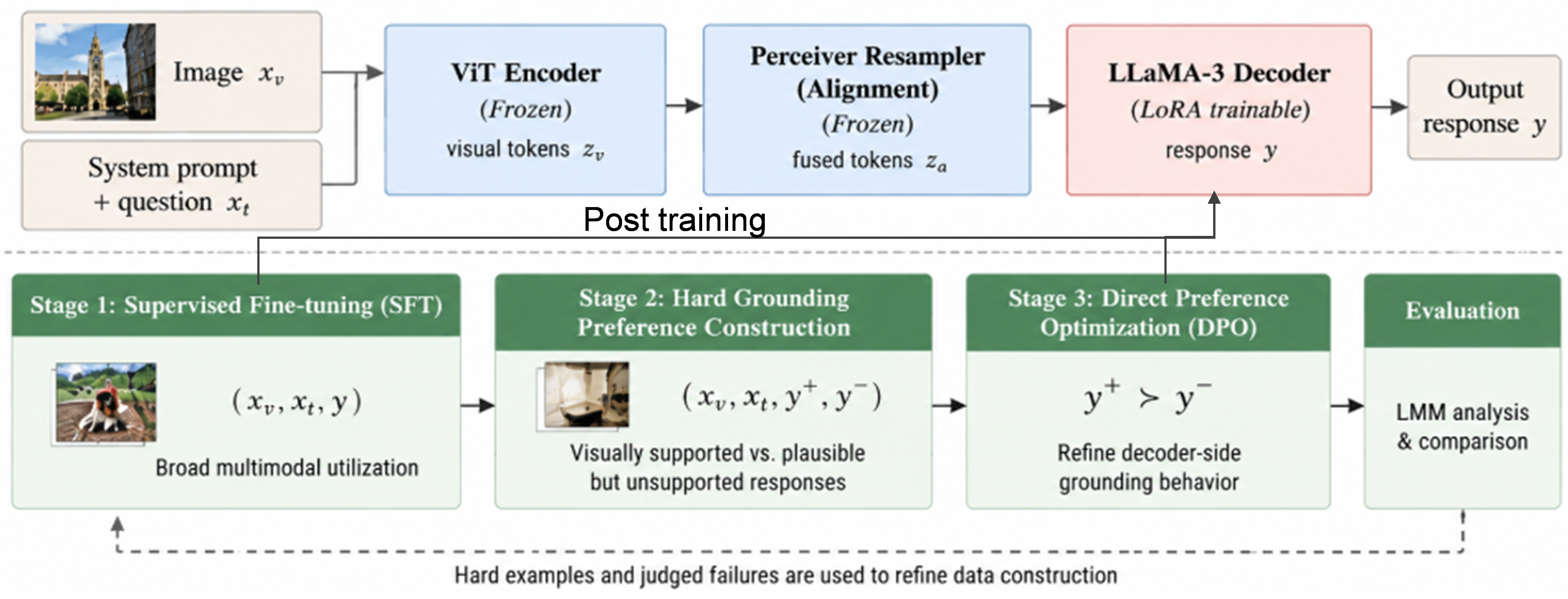}
    \caption{Fixed-visual-pathway architecture and stage-wise decoder
    post-training. The ViT encoder and Perceiver-based alignment module remain
    frozen throughout training, while Stage-1 SFT and Stage-2 DPO update only
    decoder-side LoRA parameters. Hard grounding preference pairs contrast
    visually supported responses with plausible but unsupported alternatives,
    targeting how the decoder utilizes fixed multimodal representations for
    grounded generation.}
    \label{fig:training_pipeline}
\end{figure}

Our framework does not modify the model architecture. As illustrated in
Figure~\ref{fig:training_pipeline}, the ViT encoder and Perceiver alignment
module remain frozen throughout post-training, while only LoRA parameters in
the language decoder are updated. Post-training proceeds in two stages:
large-scale multimodal SFT first establishes broad decoder-side multimodal
question answering and grounding behavior, after which targeted DPO refines
the decoder using hard grounding preference pairs. This fixed-pathway design
provides the experimental control used throughout the paper: changes from SFT
to DPO reflect decoder-side adaptation under the same visual feature
extraction and cross-modal alignment pathway.

\subsection{Hard Grounding Preference Construction}

Let $x=(x_v,x_t)$ denote an image--question pair and $y$ its grounded SFT
response. From selected SFT examples, we construct harder preference
instances $(x',y^+,y^-)$ on the same image. A generator MLLM is provided
with the original question--response pair and a target grounding behavior to
generate a harder question and candidate responses. Candidates are
subsequently adjudicated and, when necessary, manually refined. Thus,
$y^+$ and $y^-$ are often both fluent and plausible while differing primarily
in their support from visual evidence. Generation and evaluation prompt
templates are provided in the HardVQA-DPO repository.

Rather than defining hard examples solely by task category, we organize
preference construction around three complementary grounding behaviors.
\textbf{Evidence utilization} targets whether subtle but available evidence
is used, including small, partially visible, crowded, or visually confusable
objects. \textbf{Evidence calibration} targets whether response specificity
matches the available evidence: $y^+$ rejects or reduces specificity when
evidence is insufficient, but retains a specific answer when difficult yet
sufficient evidence is present. \textbf{Grounding consistency} targets
whether individual claims and reasoning remain consistent with the image,
including cases in which $y^+$ and $y^-$ reach the same final decision but
differ in their visual justification. These behaviors are not mutually
exclusive and span conventional tasks including counting, OCR, attributes,
spatial relationships, object existence, and false-premise questions.

The central design principle is to keep substantial semantic content shared
between $y^+$ and $y^-$ while preserving a grounding-relevant distinction.
A hard pair may differ only in a count, attribute, spatial relation,
object-existence claim, or reasoning step. Importantly, the construction is
bidirectional: insufficient evidence should lead to rejection or reduced
specificity, whereas difficult but sufficient evidence should retain the
supported prediction. Preference learning therefore cannot succeed merely by
making the model uniformly more conservative.

\begin{figure*}[!ht]
\centering
\setlength{\tabcolsep}{2pt}


\begin{minipage}[!ht]{0.325\textwidth}
\begin{minipage}[!ht]{0.36\linewidth}
    \vspace{0pt}
    \includegraphics[width=\linewidth]{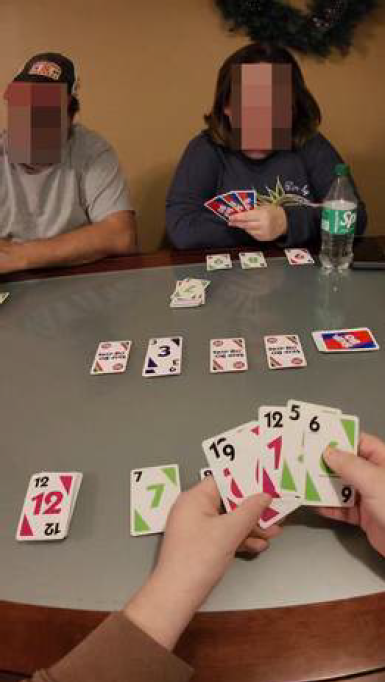}
\end{minipage}\hfill
\begin{minipage}[!ht]{0.61\linewidth}
    \vspace{0pt}\scriptsize
    \textbf{(a) Occluded evidence}\\
    \textbf{Q:} How many cards are currently in my hand?\\
    \textbf{$+$ Preferred: 5.}
    Five cards are visible, including a partially occluded fifth card.\\
    \textbf{$-$ Rejected: 4.}
    Only four cards are identified as clearly visible.
\end{minipage}
\end{minipage}
\hfill
\begin{minipage}[!ht]{0.325\textwidth}
\begin{minipage}[!ht]{0.36\linewidth}
    \vspace{0pt}
    \includegraphics[width=\linewidth]{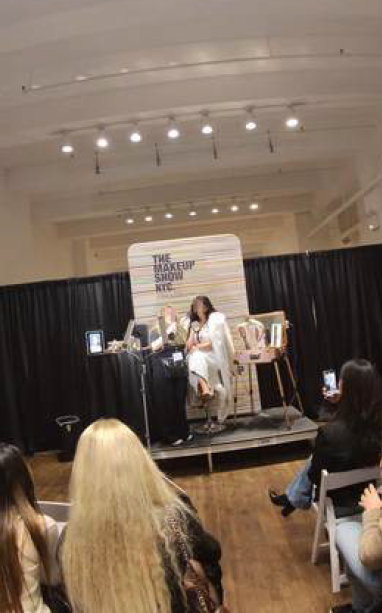}
\end{minipage}\hfill
\begin{minipage}[!ht]{0.61\linewidth}
    \vspace{0pt}\scriptsize
    \textbf{(b) Crowded instances}\\
    \textbf{Q:} How many spotlights are in the row closest to the top?\\
    \textbf{$+$ Preferred: 7.}
    Careful counting gives approximately seven fixtures despite fisheye
    distortion.\\
    \textbf{$-$ Rejected: 6.}
    The uppermost row is judged to contain exactly six visible fixtures.
\end{minipage}
\end{minipage}
\hfill
\begin{minipage}[!ht]{0.325\textwidth}
\begin{minipage}[!ht]{0.36\linewidth}
    \vspace{0pt}
    \includegraphics[width=\linewidth]{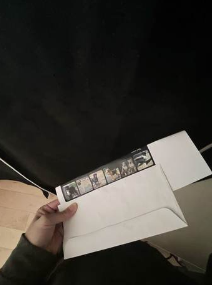}
\end{minipage}\hfill
\begin{minipage}[!ht]{0.61\linewidth}
    \vspace{0pt}\scriptsize
    \textbf{(c) Difficult but sufficient}\\
    \textbf{Q:} In the fifth photo-strip frame from the left, how many dogs
    can be seen?\\
    \textbf{$+$ Preferred: 2.}
    The fifth frame shows two dogs standing together; preceding frames show
    one dog each.\\
    \textbf{$-$ Rejected: 1.}
    The rotated, small, partially obscured strip is judged too unclear to
    confirm two dogs.
\end{minipage}
\end{minipage}

\vspace{2.5mm}


\begin{minipage}[!ht]{0.325\textwidth}
\begin{minipage}[!ht]{0.36\linewidth}
    \vspace{0pt}
    \includegraphics[width=\linewidth]{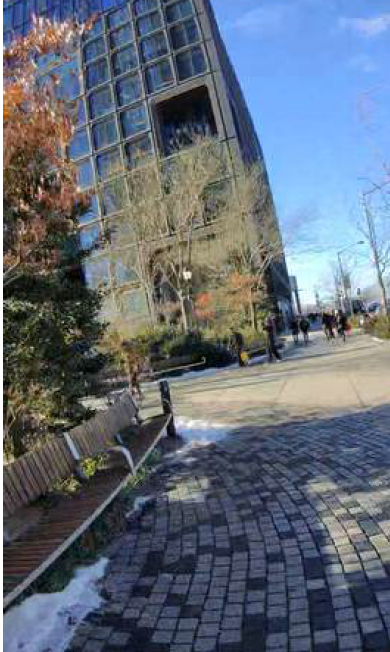}
\end{minipage}\hfill
\begin{minipage}[!ht]{0.61\linewidth}
    \vspace{0pt}\scriptsize
    \textbf{(d) Insufficient evidence}\\
    \textbf{Q:} If I am walking North, the path turns East.\\
    \textbf{$+$ Preferred: Unanswerable.}
    The tilted view provides no compass, map, or directional cue for
    determining East or West.\\
    \textbf{$-$ Rejected: No.}
    The path is asserted to curve left (West), implicitly treating image-left
    as West.
\end{minipage}
\end{minipage}
\hfill
\begin{minipage}[!ht]{0.325\textwidth}
\begin{minipage}[!ht]{0.36\linewidth}
    \vspace{0pt}
    \includegraphics[width=\linewidth]{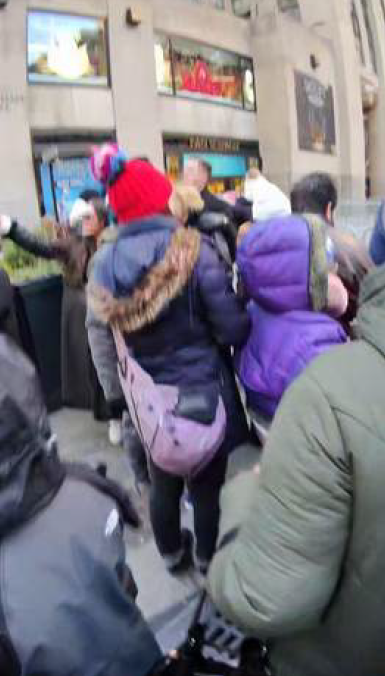}
\end{minipage}\hfill
\begin{minipage}[!ht]{0.61\linewidth}
    \vspace{0pt}\scriptsize
    \textbf{(e) False premise}\\
    \textbf{Q:} Is there a small animal, like a rat, visible peering out from
    the storm drain?\\
    \textbf{$+$ Preferred: No.}
    No animal is visible near or inside the drain grate; only feet and
    pavement are visible.\\
    \textbf{$-$ Rejected: No.}
    It instead claims that no storm-drain grate is identifiable, failing to
    ground the rejection in the visible drain.
\end{minipage}
\end{minipage}
\hfill
\begin{minipage}[!ht]{0.325\textwidth}
\begin{minipage}[!ht]{0.36\linewidth}
    \vspace{0pt}
    \includegraphics[width=\linewidth]{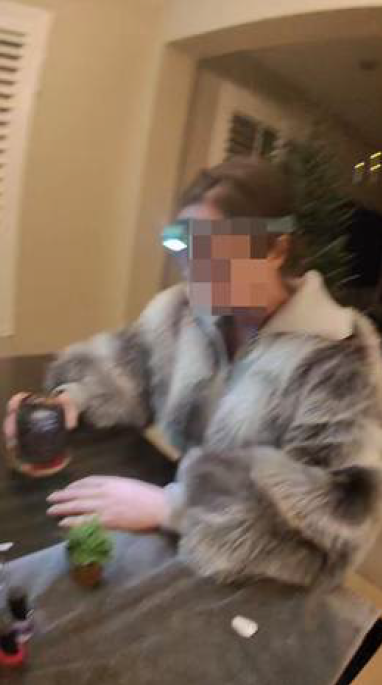}
\end{minipage}\hfill
\begin{minipage}[!ht]{0.61\linewidth}
    \vspace{0pt}\scriptsize
    \textbf{(f) Grounding consistency}\\
    \textbf{Q:} Is the person holding a smartphone in their right hand?\\
    \textbf{$+$ Preferred: No.}
    The person is holding a glass in their right hand, not a smartphone.\\
    \textbf{$-$ Rejected: No.}
    It identifies a cup/glass but incorrectly assigns it to the person's
    left hand and misinterprets the left--right relation.
\end{minipage}
\end{minipage}

\caption{
\textbf{Representative hard grounding preference pairs constructed from
WearVQA images.}
Each example pairs a hard visual question with a preferred ($+$) and rejected
($-$) response whose distinction depends on visual grounding rather than
fluency alone.
\textbf{(a--b)} Fine-grained evidence utilization requires resolving
partially occluded or crowded instances.
\textbf{(c--d)} Evidence calibration is bidirectional: difficult but
sufficient evidence should retain a specific prediction, whereas genuinely
insufficient evidence should not induce an unsupported inference.
\textbf{(e--f)} Grounding consistency distinguishes responses that may reach
the same final decision but differ in whether their supporting claims are
consistent with the image.
The examples are illustrative rather than mutually exclusive and span the
broader dimensions of evidence utilization, evidence calibration, and
grounding consistency. All images are from WearVQA.
}
\label{fig:hard_grounding_pairs}
\end{figure*}

Figure~\ref{fig:hard_grounding_pairs} illustrates how these grounding
dimensions are instantiated across different visual tasks and conditions. When both responses are visually valid, preference is instead determined by
question relevance and grounded informativeness, encouraging additional
detail when that detail is supported by the image. 

To facilitate inspection and reuse, we release \textbf{HardVQA-DPO}, a
curated and growing resource containing more than 3K hard-grounding SFT
examples and an initial set of DPO preference pairs. Additional construction
examples are provided in Appendix~\ref{app:hardvqa}.

\subsection{Hard False-Premise Construction}

We construct false-premise examples by introducing an object, attribute, or
relation that is plausible in context but unsupported by the image. Rather
than using obviously absent concepts, we select premises that are visually
confusable with nearby content because of appearance, occlusion, spatial
proximity, or scene context. For example, Figure~\ref{fig:grounding_failure_modes}(d)
asks about a bug although the visible object is not a bug, creating pressure
to complete the plausible premise with a specific insect identity.
Figure~\ref{fig:hard_grounding_pairs}(e) shows a harder case in which both
responses reject a proposed rat, but only the preferred response grounds the
rejection in the visible drain. Thus, preference is assigned not merely for
rejecting a false premise, but for matching the response and its justification
to the available visual evidence.

\subsection{Response Detail and Length Control}

A potential confound in hallucination-oriented preference construction is
response length: unsupported claims can be reduced trivially by producing
shorter or less informative answers. Our objective instead is to suppress
unsupported specificity while preserving supported detail. We therefore use
two complementary controls. First, length-aware resampling applies
exponential tilting, assigning a response of length $l$ the weight
$w(l)=\exp[\beta(l-\mu)]$, where $\mu$ is the mean response length and the
weights are capped to avoid over-amplifying extreme long-tail examples.
Second, for selected SFT examples we expand short responses with additional
visually supported attributes, spatial relationships, and contextual details.
These controls provide richer supervision for constructing $y^+$ and $y^-$
and reduce the likelihood that preference is determined merely by response
length or level of detail.

\subsection{Preference Data Quality Control}

Because hard preference pairs can differ only in fine-grained details, we use
multi-stage quality control to verify visual grounding and preference
validity. Candidate pairs are generated using one LLM and independently
evaluated by two LLM judges for whether (i) $y^+$ is supported by the image,
(ii) $y^-$ represents a valid lower-preference response, and (iii) the
preference distinction is consistent with the intended grounding difficulty.
An agentic metadata-processing stage consolidates the judgments and associated
metadata. We additionally manually review approximately 3\% of
judge-agreement samples together with judge-disagreement cases. Only examples
passing the resulting filtering process are retained for preference training.

\subsection{Stage-wise Grounded Post-Training}
\label{sec:stagewise_training}

Rather than constructing an independent preference dataset after SFT, we
derive $\mathcal{D}_2$ primarily by extending selected grounded examples from
the SFT dataset $\mathcal{D}_1$ into harder, semantically close preference
boundaries. Stage 1 performs broad multimodal SFT, updating only decoder-side
LoRA parameters while the visual encoder and alignment module remain frozen.
Let $\pi_{\mathrm{SFT}}$ denote the resulting policy. Stage 2 initializes from
$\pi_{\mathrm{SFT}}$ and applies DPO~\citep{rafailov2023direct} to
$\mathcal{D}_2$. For a preference triple $(x,y^+,y^-)$, we optimize

\begin{equation}
\small
\mathcal{L}_{\mathrm{DPO}}
=
-\log \sigma\!\left(
\beta
\left[
\log\frac{\pi_\theta(y^+\mid x)}
         {\pi_{\mathrm{SFT}}(y^+\mid x)}
-
\log\frac{\pi_\theta(y^-\mid x)}
         {\pi_{\mathrm{SFT}}(y^-\mid x)}
\right]
\right),
\label{eq:grounded_dpo}
\end{equation}

where $\pi_{\mathrm{SFT}}$ is the fixed DPO reference policy and $\beta$
controls the preference scale. The DPO objective itself follows standard
DPO; our distinction lies in how preference data are coupled to the SFT
stage. Selected grounded SFT cases are transformed into local contrasts
between visually supported and plausible but unsupported continuations,
concentrating second-stage supervision on difficult grounding boundaries
while retaining the SFT checkpoint as both initialization and reference
policy. The resulting procedure is \textbf{broad multimodal SFT
$\rightarrow$ hard grounding preference construction $\rightarrow$
grounding-focused DPO}.

\section{Experimental Results and Analysis}

We evaluate the proposed approach along two complementary dimensions:
(i) whether hard grounding preference supervision improves the Stage-1 SFT
model on challenging multimodal responses, and (ii) whether this targeted
decoder-side refinement preserves broader multimodal capabilities. For general
capability, we evaluate DocVQA \citep{mathew2021docvqa}, TextVQA \citep{singh2019towards}, MMBench
\citep{liu2023mmbench}, and VQAv2 \citep{antol2015vqa}. For grounding-sensitive
evaluation, we compare Stage-1 SFT and Stage-2 DPO responses on approximately
839 diverse multimodal examples using three independent LLM judges: GPT-4o,
Gemini-2.5-Flash, and LLaMA-4-Maverick. The judges consider visual grounding,
correctness, informativeness, and unsupported content
\citep{zheng2023judging}. We additionally compare against Qwen2.5-VL-7B and
InternVL3-8B on standard benchmarks and use selected proprietary-model outputs
only as qualitative cross-model case studies.

\subsection{Quantitative Results}

\textbf{Performance on Standard Multimodal Benchmarks:} Table~\ref{tab:sft_benchmark} examines whether targeted preference refinement
preserves the broader capabilities acquired during SFT. The controlled
comparison of primary interest is between the Stage-1 SFT and Stage-2 DPO
checkpoints, which share the same frozen visual pathway and differ only in
decoder-side post-training.

\begin{table}[!ht]
\centering
\caption{Performance comparison on standard multimodal benchmarks.}
\label{tab:sft_benchmark}
\resizebox{0.7\columnwidth}{!}{
\begin{tabular}{lcccc}
\hline
Model & DocVQA & TextVQA & MMBench & VQAv2 \\
\hline
Qwen2.5-VL-7B  & \textbf{87.08} & 72.77 & \textbf{82.60} & 67.81 \\
InternVL3-8B   & 75.02 & 63.35 & 82.49 & 61.86 \\
LLaMA-3 MM SFT & 81.5 & 80.5 & 78.0 & 81.5 \\
LLaMA-3 MM DPO & 83.0 & \textbf{81.6} & 80.1 & \textbf{82.2} \\
\hline
\end{tabular}
}
\end{table}

\textbf{Evaluation details:} Validation splits are used because official test-set annotations are not
publicly available. Evaluation sizes are DocVQA validation (5,349), TextVQA validation
(5,000), MMBench development (dev, 4,329), and VQAv2 validation (47,206 questions). DocVQA is evaluated using ExactMatch-Norm for all models to maintain a consistent
metric rather than ANLS. Complete evaluation scripts and benchmark
implementations are provided in the anonymous repository.

The DPO checkpoint improves over its SFT initialization on all four
benchmarks: 81.5$\rightarrow$83.0 on DocVQA,
80.5$\rightarrow$81.6 on TextVQA, 78.0$\rightarrow$80.1 on MMBench, and
81.5$\rightarrow$82.2 on VQAv2. Thus, refinement concentrated on difficult
grounding examples does not appear to trade away broader multimodal
capabilities. Qwen2.5-VL-7B remains stronger on DocVQA and MMBench, consistent
with its strong document/OCR capabilities and MMBench's constrained
multiple-choice evaluation, whereas our DPO checkpoint is stronger on the
more open-ended TextVQA and VQAv2 under this protocol. Since the external
models differ in architecture and training, these comparisons are contextual;
the controlled result is the consistent DPO improvement over SFT.

\textbf{Pairwise Grounding Evaluation:} We next directly compare Stage-2 DPO with its Stage-1 SFT checkpoint on
approximately 839 multimodal examples. For each image--query pair, responses
from the two checkpoints are evaluated independently by GPT-4o,
Gemini-2.5-Flash, and LLaMA-4-Maverick.

\begin{table}[!ht]
\caption{Pairwise comparison of Stage-1 SFT and Stage-2 DPO responses across
three independent LLM judges on approximately 839 multimodal examples.}
\label{tab:dpo_pairwise}
\centering
\resizebox{0.7\columnwidth}{!}{
\begin{tabular}{lccc}
\hline
Metric & GPT-4o & Gemini-2.5-Flash & LLaMA-4-Maverick \\
\hline
LLaMA-3 MM SFT Wins & 207 (24.7\%) & 256 (30.5\%) & 265 (31.6\%) \\
LLaMA-3 MM DPO Wins & 276 (32.9\%) & 316 (37.7\%) & 317 (37.8\%) \\
Equivalent (TIE/C) & 294 (35.0\%) & 221 (26.3\%) & 220 (26.2\%) \\
Both Wrong (D) & 49 (5.8\%) & 20 (2.4\%) & 19 (2.3\%) \\
Error Rate & 13 (1.5\%) & 26 (3.1\%) & 18 (2.1\%) \\
$\Delta$ Win Rate (pp) & +8.2 & +7.2 & +6.2 \\
\hline
\end{tabular}
}
\end{table}

All three judges favor DPO more often than SFT, with DPO-minus-SFT win-rate
differences of +8.2, +7.2, and +6.2 percentage points, respectively. The
consistent direction across three evaluators reduces dependence on any
single judge.

\textbf{Human validation:} We manually review approximately 3\% of judge-agreement cases together with
judge-disagreement cases. This audit checks the reliability of the automated
pairwise judgments and should not be interpreted as an independent
large-scale human evaluation.

Together, the benchmark and pairwise results provide complementary evidence
that targeted decoder-side preference refinement improves judged grounding
and response quality without degrading broader multimodal performance. These
results use approximately 50k--120k targeted preference examples. Because
preference difficulty and dataset size are not independently varied, the
current experiments do not isolate the contribution of hardness from that of
data scale.

\begin{figure}[!ht]
    \centering
    \includegraphics[width=\linewidth]{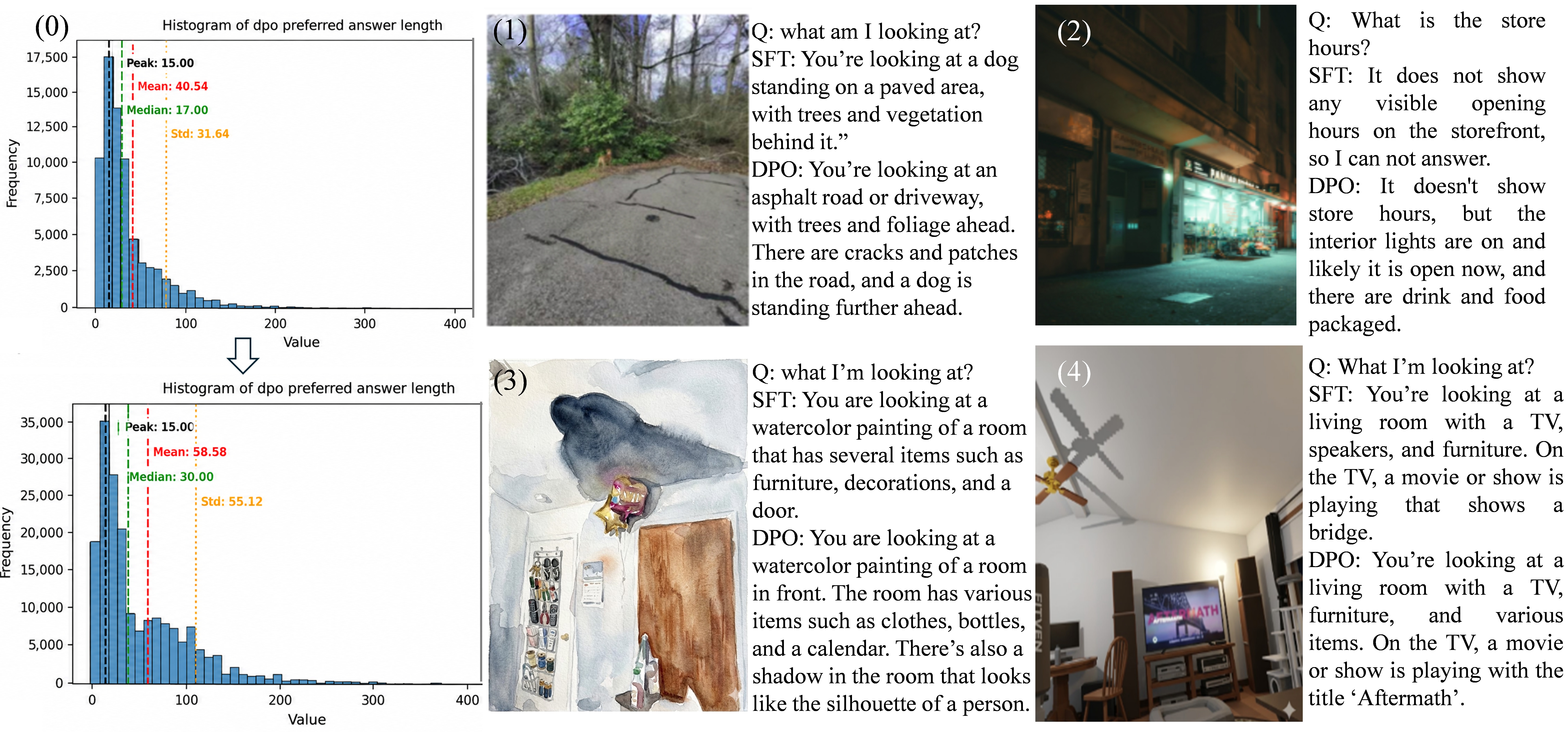}
\caption{\textbf{Grounded informativeness and response-detail control.}
\textbf{(0)} Preferred-response length distributions before and after
length-aware resampling and response-detail expansion, which increase
representation of informative longer responses rather than favoring
uniformly short answers. \textbf{(1--4)} Representative
SFT$\rightarrow$DPO comparisons showing that preference optimization can
retain or add visually supported, query-relevant details, including
fine-grained scene attributes, spatial relationships, and contextual
information. Together, these results complement the hallucination-reduction
examples by showing that improved grounding does not arise merely from
shortening responses or suppressing specificity indiscriminately. The first
image was captured by the author; the second is from Krüger (2024,
Unsplash); the third and fourth are AI-generated.}
\label{fig:grounded_informativeness}
\end{figure}

\textbf{Grounded Informativeness and Response Detail:} Reducing hallucination should not come at the cost of uniformly shorter or
less informative responses. We therefore apply length-aware resampling to
increase the representation of informative longer responses. For a response
of length $l$, we assign the sampling weight
$w(l)=\exp[\beta(l-\mu)]$, where $\mu$ is the mean response length and
$\beta$ controls the strength of length-aware reweighting; the weights are
capped to avoid over-amplifying extreme long-tail responses.
Figure~\ref{fig:grounded_informativeness}(0) shows the response-length
distribution before and after this procedure. In addition, selected short
responses are expanded with visually supported attributes, spatial
relationships, and contextual details.

Figures~\ref{fig:grounded_informativeness}(1--4) provide representative
SFT$\rightarrow$DPO comparisons. DPO can retain or add query-relevant detail
when supported by the image, including fine-grained scene attributes,
spatial relationships, and visible contextual information. Together, the
distributional and qualitative results show that the observed reduction in
unsupported specificity is not explained simply by shorter or uniformly more
conservative generation: preference supervision can suppress unsupported
claims while preserving or increasing supported detail.

\subsection{How Grounding Behavior Changes after DPO}

The quantitative results establish an aggregate improvement; we next examine
\emph{how} behavior changes under the same frozen visual pathway in terms of
evidence utilization, calibration, and grounding consistency. The desired
behavior is bidirectional: suppress unsupported specificity when evidence is
insufficient while preserving supported specificity when useful visual
evidence is available.

\begin{figure}[!ht]
    \centering
    \includegraphics[width=1\linewidth]{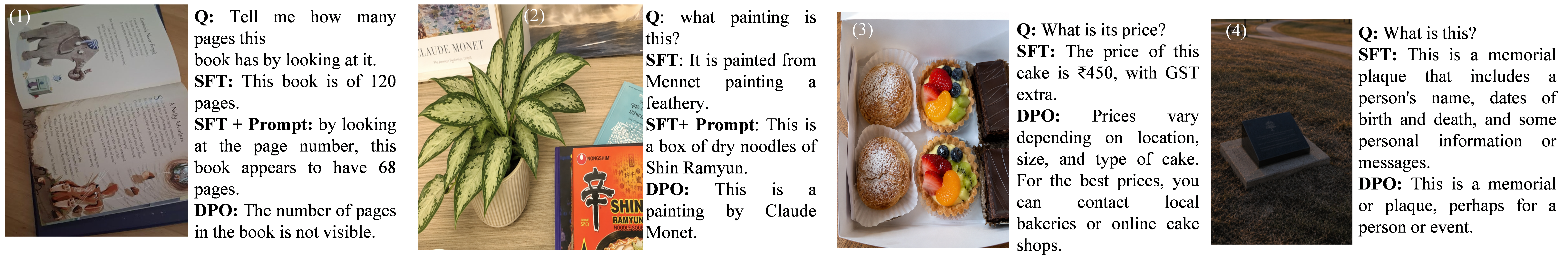}
    \caption{Representative SFT$\rightarrow$DPO changes in visually grounded
    generation. Across the four examples, DPO reduces unsupported commitments
    made by the Stage-1 SFT model, including an unverifiable book-page count,
    an unreliable visual identification, an unsupported product price, and
    unsupported details about a memorial. These examples illustrate
    improvements across numerical, recognition, and semantic grounding
    failures. The first image was captured by the author; the second is from
    Krüger (2024, Unsplash); the third and fourth are AI-generated.}
    \label{fig:dpo_behavior}
\end{figure}

\textbf{Supported versus Unsupported Specificity:} Figure~\ref{fig:dpo_behavior} presents representative SFT$\rightarrow$DPO
changes selected from the larger qualitative analysis. When the image does
not provide sufficient evidence---for example, when the total number of book
pages or a fixed product price is not visible---the DPO model avoids specific
unsupported values produced by SFT. The examples also extend beyond numerical
claims: DPO corrects an unreliable visual interpretation in the painting
example and avoids unsupported identity and biographical details in the
memorial example. Together, these cases show that hard grounding preference
optimization can reduce plausible-but-unsupported commitments across
numerical, recognition, and semantic grounding failures rather than merely
altering response length or style.

\textbf{Remaining Failure Modes:} Hard grounding preference supervision does not eliminate hallucination;
residual errors include difficult spatial reasoning and occasional unsupported
object insertion. Such errors do not by themselves reveal whether the limiting
factor is the fixed representation or decoder use of that representation.
Accordingly, our results show that decoder-side adaptation is sufficient to
improve a meaningful subset of grounding failures, not that the decoder is the
sole source of multimodal hallucination.

\begin{figure}[!ht]
    \centering
    \includegraphics[width=\linewidth]{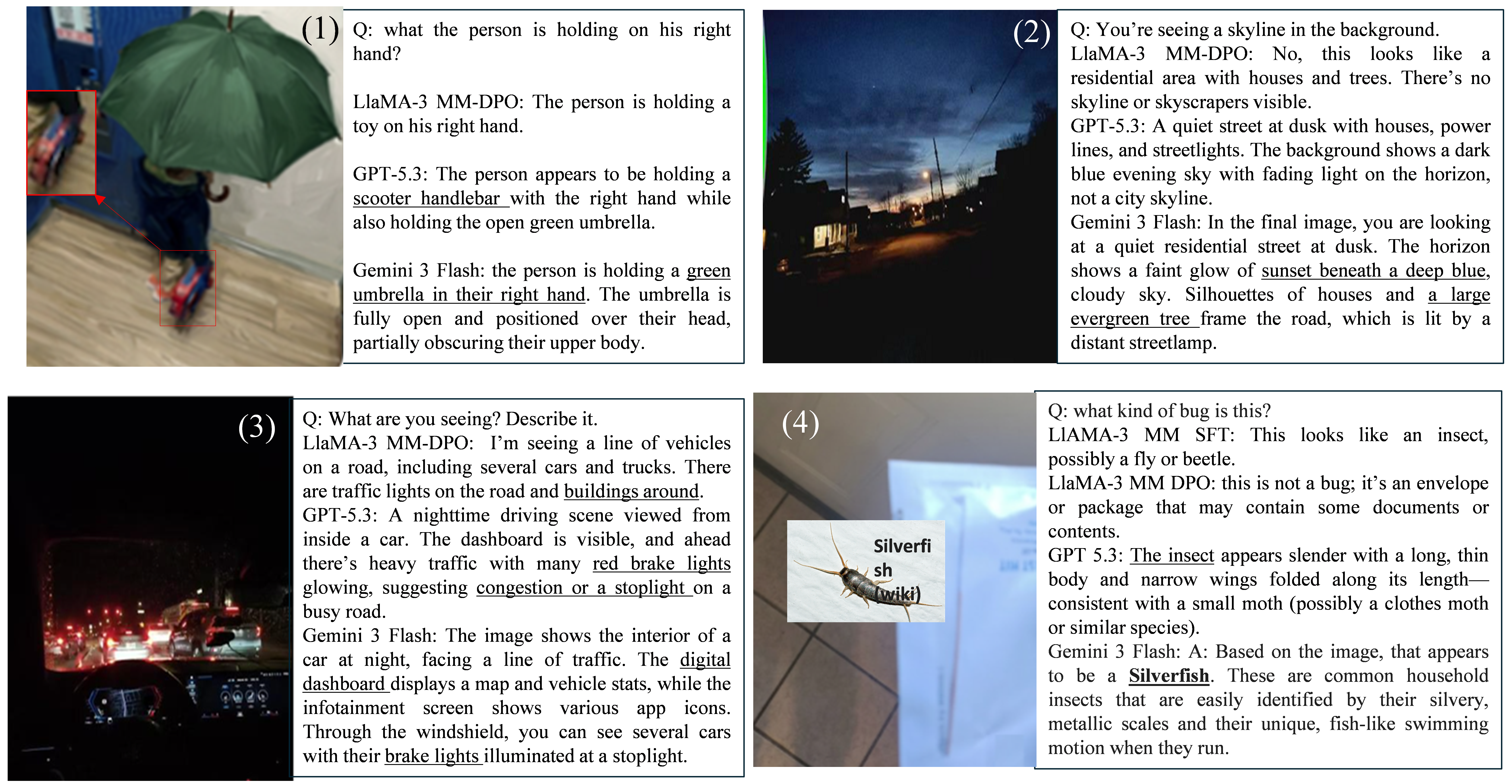}
\caption{Cross-model qualitative examples spanning complementary grounding
challenges: \textbf{(1)} physical and spatial grounding under partial
visibility, \textbf{(2)} interpretation under severely limited visual
evidence, \textbf{(3)} fine-grained grounding in a crowded traffic scene,
and \textbf{(4)} false-premise grounding. Across these examples, the models
differ in their use of available visual evidence and in the extent to which
plausible but unsupported interpretations are introduced or elaborated.
These cases illustrate grounding challenges involving evidence utilization,
evidence calibration, spatial reasoning, and unsupported generation; the
comparison is diagnostic rather than a ranking of models.}
\label{fig:hallu_cross}
\end{figure}

\section{Conclusion}
We study multimodal hallucination through hard grounding preference
supervision, where preferred and rejected responses are semantically close
but differ in their support from visual evidence. Under a frozen visual
encoder and alignment pathway, SFT establishes broad decoder-side multimodal
behavior, while targeted DPO refines evidence utilization, calibration, and
grounding consistency. This design targets both unsupported specificity and
unnecessary rejection rather than simply encouraging more conservative
generation.

The DPO model improves over its SFT initialization across DocVQA, TextVQA,
MMBench, and VQAv2 and achieves 6.2--8.2 percentage-point higher pairwise win
rates across three independent judges on approximately 839 additional
multimodal examples. These results show that decoder-side adaptation alone can
improve a meaningful subset of grounding failures when useful visual evidence
is available through the fixed multimodal representation. We release
HardVQA-DPO to support further study of hard grounding supervision and
visually grounded multimodal generation.





\section*{Generative AI Usage Disclosure}

Generative AI tools were used in this work for synthetic data construction
and model-based evaluation, as described in the corresponding methodology
and evaluation sections. Generative AI tools were also used to assist with
literature discovery and to partially draft, edit, and improve the clarity
of portions of the manuscript. All AI-assisted content and references were
reviewed and verified by the authors, who take full responsibility for the
final content of the paper.

\bibliography{iclr2027_conference_revised}
\bibliographystyle{iclr2027_conference}

\section*{Appendix}
\appendix


\section{HardVQA-DPO Construction Examples}
\label{app:hardvqa}

The following examples provide representative end-to-end illustrations of
the HardVQA-DPO construction process, including the original SFT question
and response, the harder grounding-targeted DPO question, the corresponding
preferred and rejected responses with rationales, and human-review metadata.
They cover multiple grounding behaviors discussed in the main paper,
including under-utilization of subtle visual evidence, unnecessary rejection,
unsupported inference, false-premise handling, and grounding-inconsistent
reasoning. These behaviors are not mutually exclusive and occur across
conventional tasks such as counting, object existence, attributes, and
spatial reasoning.

A subset of these examples is also presented in condensed form in
Figure~\ref{fig:hard_grounding_pairs} of the main paper, where only the
grounding-relevant DPO preference pairs are shown for clarity. We retain the
complete examples here to document the original SFT question--answer pairs,
full preference rationales, and associated human-review metadata, providing
additional transparency into the construction and quality-control process.

All images shown in this appendix are sourced from the publicly available
WearVQA dataset \citep{chang2025wearvqa}, hosted on Hugging Face under the
Apache-2.0 license. WearVQA contains egocentric real-world images designed
for visual question answering in wearable-device scenarios.

\clearpage
\makeatletter
\let\oldmakeLineNumber\makeLineNumber
\makeatother

\includepdf[
  pages=-,
  fitpaper=true,
  pagecommand={\thispagestyle{plain}}
]{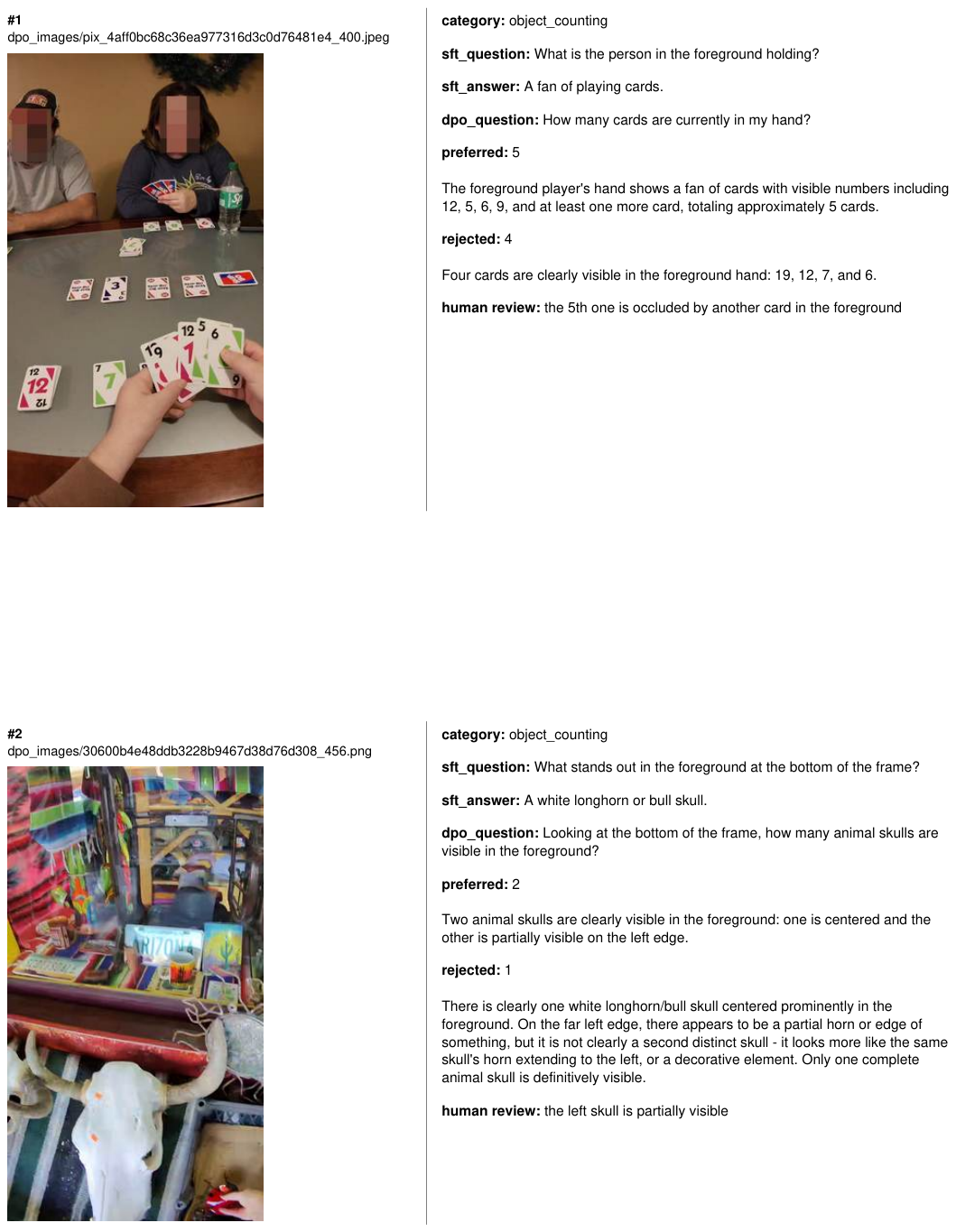}

\makeatletter
\let\makeLineNumber\oldmakeLineNumber
\makeatother
\clearpage


\section{Additional Cross-Model Qualitative Analysis}
\label{app:crossmodel}

This appendix provides additional analysis of the cross-model examples
introduced in the main text. The purpose is not to establish a statistically
representative ranking of model families, but to examine how different
systems respond when visual evidence is ambiguous, incomplete, or
insufficient for a unique answer. These cases complement the controlled
SFT--DPO comparison by illustrating how similar grounding difficulties can
appear across models with different architectures and training procedures.

\subsection{Detailed Analysis of the False-Premise Example}
\label{app:bug_analysis}

We first revisit the false-premise bug example discussed in the main text.
There, the example illustrates a general cross-model behavior: when a
question introduces an unsupported premise, different MLLMs can make
different plausible commitments and subsequently elaborate on them. Here we
examine the individual responses in greater detail from the perspectives of
evidence utilization, response specificity, and premise following.


The responses reveal two related effects. First, different models assign
different specific interpretations to the same ambiguous visual evidence.
The disagreement itself does not identify the source of the error, but it is
consistent with model-specific linguistic and contextual priors influencing
generation when the image does not strongly determine the answer. Second,
after making an initial identification, a model can generate additional
details that are coherent with that identification even when the initial
premise is unsupported. Thus, the problem is not simply verbosity: an
autoregressive continuation can be internally coherent while remaining
grounded in an unsupported initial commitment.

The DPO-trained model instead rejects the assumed object in this example.
Because its ViT encoder and Perceiver alignment module are unchanged from the
SFT checkpoint, the SFT--DPO behavioral difference is consistent with
decoder-side post-training changing how the fixed multimodal evidence is
used. This example does not establish that the original failure arose
exclusively in the decoder, and the responses of external models cannot
localize their failures to a particular component because their multimodal
architectures differ.

\subsection{Detailed Cross-Model Analysis of Grounding Behavior}

The remaining examples introduced in the main text span unsupported object
inference, spatial reasoning under partial occlusion, ambiguous visual
conditions, OCR and fine-grained recognition, and contextual interpretations
that extend beyond directly observable evidence. Here we examine these cases
through the three complementary grounding dimensions used throughout the
paper: evidence utilization, evidence calibration, and grounding consistency.

\textbf{Evidence utilization:}
When relevant evidence is small, partially visible, crowded, or otherwise
difficult to perceive, responses can fail by under-utilizing information that
is nevertheless present in the image. This can appear as an incorrect count,
a missed object or attribute, or an unnecessarily generic answer. Importantly,
the opposite behavior is also possible: a model may reject a question even
when subtle but sufficient evidence supports a specific answer. These cases
motivate preference pairs that distinguish difficult evidence from genuinely
insufficient evidence.

\textbf{Evidence calibration:}
When the image is underdetermined, models differ in how much specificity they
introduce. Some responses provide plausible prices, object identities,
attributes, spatial relations, or contextual interpretations that cannot be
verified from the image. Other responses appropriately reduce specificity or
state that the required information is unavailable. Greater detail is not
itself a grounding failure: several examples contain richer responses that
remain consistent with the image. The relevant distinction is whether the
specificity of the response is supported by the available evidence.

\textbf{Grounding consistency:}
Grounding errors can also occur within an otherwise plausible explanation.
In some cases, two responses reach the same final decision while differing in
whether the supporting rationale is visually correct. This motivates
evaluating grounding at the level of individual claims and reasoning steps,
rather than relying only on final-answer accuracy.

These behaviors should be interpreted as qualitative case studies rather
than model-level performance estimates. They illustrate that the
grounding--informativeness tension motivating HardVQA-DPO is not restricted
to a single task category or a single model family.

\subsection{Summary of Observed Grounding Behaviors}

Table~\ref{tab:cross_model_qualitative} summarizes the grounding behaviors
represented by the selected examples. The examples are not sampled as a
statistically representative benchmark and therefore should not be
interpreted as estimates of overall model performance.

\begin{table}[!ht]
\centering
\small
\renewcommand{\arraystretch}{1.15}
\caption{Grounding behaviors represented in selected qualitative examples.}
\label{tab:cross_model_qualitative}
\resizebox{\columnwidth}{!}{
\begin{tabular}{p{3.0cm} p{4.4cm} p{5.2cm}}
\hline
\textbf{Behavior} &
\textbf{Visual condition} &
\textbf{Representative response difference} \\
\hline

Evidence under-utilization
& Small, occluded, crowded, or fine-grained evidence
& Available visual evidence is missed, producing an incorrect or
under-specified response \\

Unnecessary rejection
& Difficult but sufficient evidence
& The response declines to answer even though the image supports a specific
conclusion \\

Unsupported inference
& Ambiguous or insufficient evidence
& A plausible object, attribute, count, price, relation, or interpretation is
introduced without sufficient visual support \\

False-premise following
& Query assumes an unsupported object, attribute, or relation
& The premise is accepted and subsequently elaborated rather than rejected or
qualified \\

Grounding inconsistency
& Multi-step or explanatory response
& The final decision can be correct while intermediate claims or justification
conflict with visible evidence \\

\hline
\end{tabular}
}
\end{table}

\subsection{Cross-Model Qualitative Examples and Analysis}

The complete set of qualitative cross-model examples is provided on the
following pages. Comparisons involving proprietary models reflect the API
versions available at the time of evaluation and may change as these systems
are updated. No external OCR or computer-vision toolkits were added to the
evaluated systems during testing.

\clearpage

\makeatletter
\let\oldmakeLineNumber\makeLineNumber
\makeatother

\includepdf[
  pages=-,
  fitpaper=true,
  pagecommand={\thispagestyle{empty}}
]{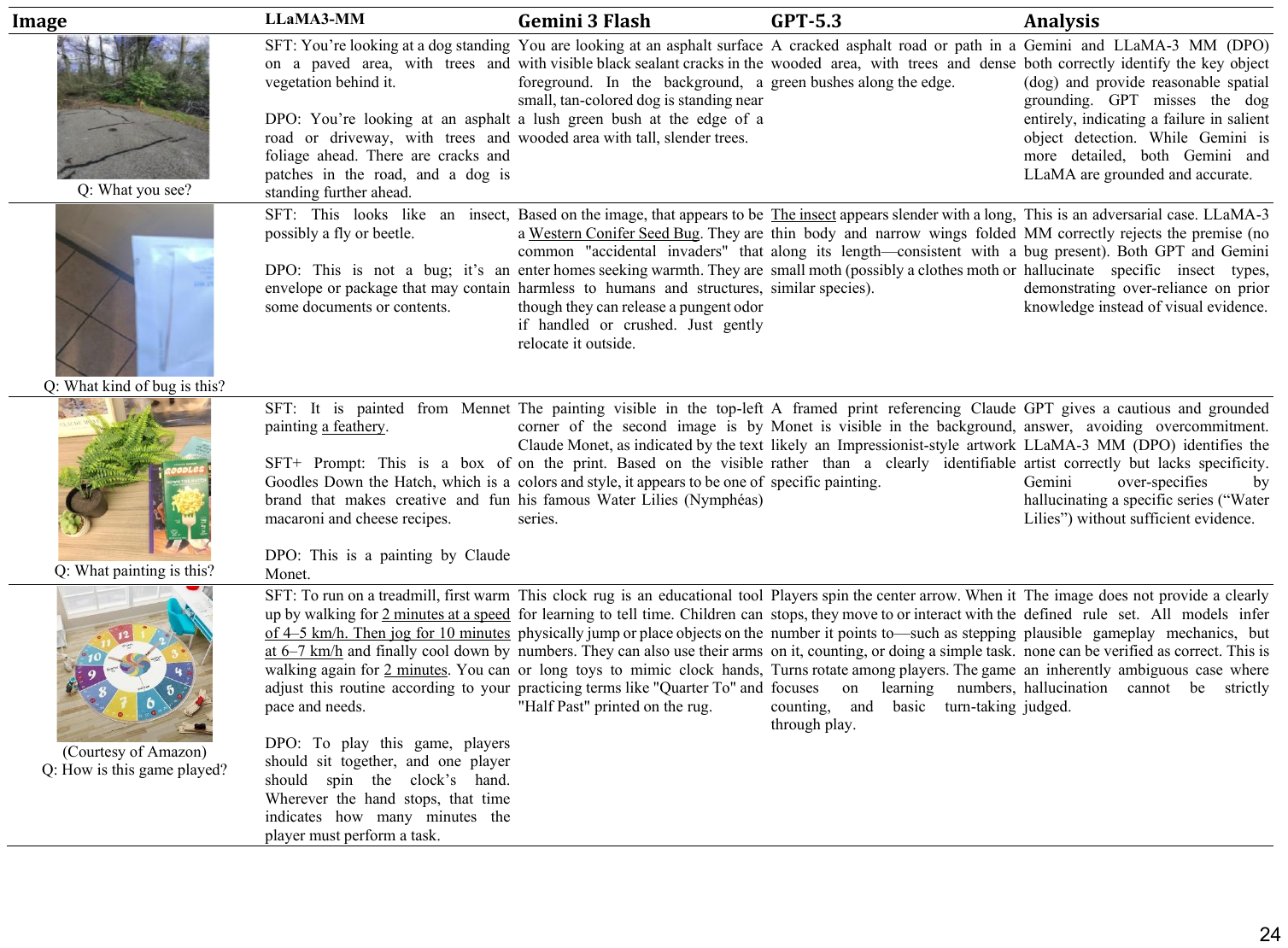}

\makeatletter
\let\makeLineNumber\oldmakeLineNumber
\makeatother

\clearpage


\section{Additional Analysis}

\subsection{Data Scale and Preference Refinement}

Stage-2 training uses approximately 50k--120k targeted preference examples,
substantially fewer than the supervised examples used in Stage 1. Across the
evaluated settings, the DPO-trained model consistently receives higher
pairwise preference rates than its SFT initialization.

The preference data concentrate on difficult grounding boundaries rather
than a single task category. They include under-utilization of subtle visual
evidence, unnecessary rejection, unsupported inference, false-premise
following, and grounding-inconsistent reasoning across counting, OCR,
attributes, spatial relations, and object-existence questions. Preferred and
rejected responses are constructed to share substantial semantic content so
that their distinction depends primarily on fine-grained grounding rather
than easily separable differences in linguistic quality.

The experiments show that targeted preference refinement provides an
effective training signal at this scale. However, preference difficulty and
dataset size are not independently varied. We therefore do not attribute the
observed gains specifically to hard-example selection rather than additional
targeted preference data more generally. A matched-scale comparison across
preference-construction strategies would be required to isolate these
effects.

\subsection{Supported and Unsupported Specificity}

A central objective of our construction is to avoid conflating improved
grounding with generic conservatism. More descriptive responses contain more
objects, attributes, relations, and contextual claims and therefore create
additional opportunities for unsupported generation. However, simply
shortening responses or rejecting difficult questions would also discard
useful information.

Our preference construction consequently operates in both directions. When
the image is insufficient to support a specific conclusion, the preferred
response avoids or qualifies unsupported detail. When subtle but sufficient
evidence is available, the preferred response retains the specific answer
and can include additional supported information. In the qualitative
examples, DPO suppresses unsupported page counts, prices, and premise-driven
interpretations while recovering additional supported OCR content and visual
details in other cases.

These opposite-direction changes are important because they are inconsistent
with a simple explanation in which DPO reduces hallucination merely by
making responses uniformly shorter or more conservative. Instead, the
observed behavior is consistent with better matching response specificity to
the available visual evidence. We do not interpret these selected examples
as a formal measurement of the detail--grounding trade-off.

\subsection{Limitations and Scope of Decoder-Side Adaptation}

\begin{figure}[!ht]
    \centering
    \includegraphics[width=\linewidth]
    {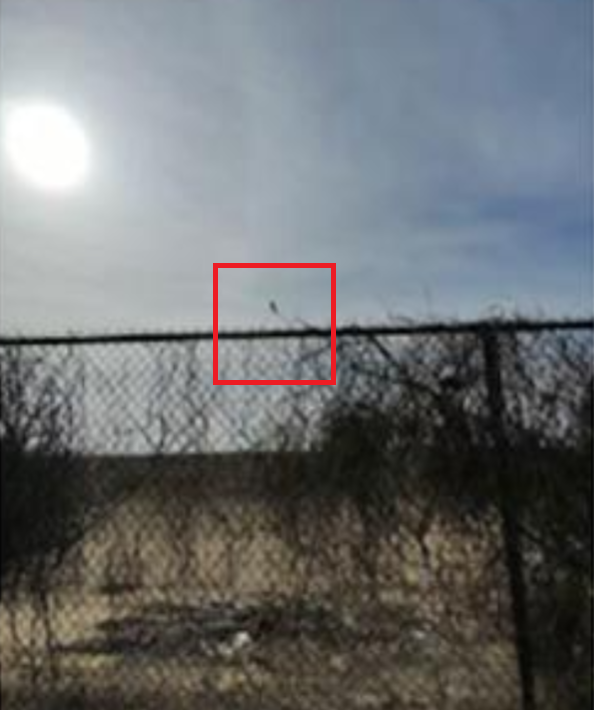}

    \caption{\textbf{Illustrative residual hallucination after DPO.}
The DPO-trained model introduces an unsupported bird on the fence,
demonstrating that hard grounding preference supervision reduces but does not eliminate plausible unsupported generation.}
    \label{fig:remaining_failures}
\end{figure}

Hard grounding preference supervision reduces but does not eliminate
unsupported generation. Figure~\ref{fig:remaining_failures} provides one
illustrative residual case in which the DPO-trained model introduces a bird
sitting on the fence despite insufficient visual support for that object.
This example is not intended to characterize the full space of residual
errors, but demonstrates that plausible unsupported content can persist after
preference optimization.

Because the ViT encoder and Perceiver alignment module remain frozen,
decoder-side post-training can improve grounding when task-relevant
information is available through the fixed multimodal representation but is
under-utilized or incorrectly interpreted by the decoder. If required visual
information is absent from that representation, decoder-side adaptation cannot
reliably recover it. The residual example does not distinguish these cases;
our controlled SFT--DPO comparison instead establishes decoder-side
sufficiency for correcting a subset of grounding failures.

\end{document}